\documentclass[runningheads]{llncs}
\usepackage[T1]{fontenc}
\usepackage{lmodern}
\usepackage{graphicx}
\usepackage{longtable}
\usepackage{amsmath}
\usepackage{dsfont}
\usepackage{amssymb}
\usepackage{url}
\usepackage{fvextra}
 \usepackage{booktabs}
 \usepackage{array}
\usepackage{xcolor}
\usepackage{comment}
\usepackage{hyperref}
\usepackage{eso-pic}

\newcommand{\springerReference}{%
  Sousa M., Pinto L. R., Matos-Carvalho J. P.
  Multi-Tool Selection for Robotic Systems: Comparing Semantic Retrieval and LLM.
  Author manuscript for the CAIESA Workshop at ESWeek 2026.
  Proceedings to be published by Springer in the IFIP Advances in Information
  and Communication Technology (AICT) series.%
}
\newcommand{\springerDOI}{} 
\newcommand{\springerNotice}{%
  \begingroup
  \setlength{\fboxsep}{6pt}%
  \setlength{\fboxrule}{0.4pt}%
  \fbox{\begin{minipage}{\dimexpr0.8\paperwidth-2\fboxsep-2\fboxrule\relax}
    \sffamily\footnotesize\raggedright
    \springerReference
    \ifx\springerDOI\empty\else
      \par\href{https://doi.org/\springerDOI}{https://doi.org/\springerDOI}%
    \fi
  \end{minipage}}%
  \endgroup
}
\AddToShipoutPictureFG*{%
  \AtPageUpperLeft{%
    \put(\LenToUnit{0.1\paperwidth},-\LenToUnit{24pt}){%
      \makebox(0,0)[lt]{\springerNotice}%
    }%
  }%
}

\newcommand{\equref}[1]{Eq.~\ref{eq:#1}}

\begin{document}

\title{Multi-Tool Selection for Robotic Systems: Comparing Semantic Retrieval and LLM}
\titlerunning{Multi-Tool Selection for Robotic Systems}

\author{Miguel Sousa\inst{1} \and Luis Ramos Pinto\inst{1} \and João P. Matos-Carvalho\inst{1}}
\authorrunning{M. Sousa, L. R Pinto, and J. P. Matos-Carvalho}
\institute{\textsuperscript{1}\enspace LASIGE, Faculdade de Ciências, Universidade de Lisboa, Portugal}

\maketitle

\begin{abstract}
Natural-language interfaces allow users to command robotic systems without interacting directly with mission planners or robotics middleware. Their reliable use depends on robot tool routing, defined here as the selection of the smallest complete subset of predefined executable capabilities required to satisfy a request. 
This task is particularly difficult for compound instructions, where incomplete selection may omit necessary sensing or monitoring capabilities, while excessive selection may expose irrelevant tools. We compare four routing strategies: three progressively refined ChromaDB-based methods -- fixed-threshold retrieval, relative-distance retrieval, and clause-aware aggregated retrieval -- and a cloud-hosted 120-billion-parameter language-model baseline. The methods are evaluated on 164 manually annotated prompts organised into three complexity levels, from direct single-tool requests to compound requests requiring at least three tools. Using exact-match accuracy, completeness, precision, a completeness-weighted composite score, and routing latency, CDB-v3 achieves the best macro-average score among the retrieval-only methods, with $0.7822$,
compared with $0.8869$ for the large-model baseline. Although the LLM provides higher routing quality, CDB-v3 exhibits substantially lower latency and can be deployed locally without requiring cloud connectivity. These results identify
a quality--latency trade-off between direct large-model selection and clause-aware semantic retrieval for privacy-sensitive, connectivity-constrained, and resource-constrained robotic systems.
\end{abstract}

\section{Introduction}

Natural-language interfaces are becoming an attractive way to interact with
robotic systems. Instead of invoking commands through mission planners,
terminals, or custom graphical interfaces, users can express requests directly
in natural language, such as ``land the drone'', ``start mapping the area'', or
``report the current battery level''. Large language models (LLMs) make such
interfaces increasingly practical, but a critical challenge remains: mapping a
user request to the correct robot capability or set of capabilities.

In robotics, capability selection is not merely a usability problem. A routing error may activate an incorrect process, produce misleading information, or issue an unintended command to the robot. As the number of available robot functions grows, reliable intent-to-capability matching becomes a fundamental requirement for safe and effective natural-language control. Traditionally, this decision logic is encoded through behaviour trees, state machines, launch configurations, or manually designed if--then rules that determine which robot capability should be activated under each condition. These approaches remain effective for predefined and well-structured workflows, but they become increasingly difficult to design, extend, and maintain as the number of capabilities, mission types, operating conditions, and possible user requests grows. Conversational interfaces shift part of this burden from handcrafted decision logic to runtime capability selection, making the routing layer a critical component of the overall system.

This challenge is particularly relevant in ROS-based systems, where a single robot or fleet may expose dozens of processes for navigation, mapping, monitoring, mission execution, telemetry collection, payload operation, and system management. A request may require one capability or a coordinated set of capabilities. For example, the instruction ``monitor the battery and land when it drops below 25\%'' requires both a telemetry capability and an
actuation capability. The routing layer must therefore determine not only which capabilities are semantically relevant, but also whether several tools
must be selected to satisfy all parts of the request.

Large LLMs can perform this selection directly by receiving the complete tool catalog and reasoning over the user request. However, dependence on a cloud-hosted model may be impractical in robotic deployments with intermittent
connectivity, limited bandwidth, high communication latency, or strict
confidentiality and data-sovereignty requirements. Robot telemetry, mission
information, maps, and operational commands may be unsuitable for transmission
to external infrastructure. Even when a large model can be deployed locally,
its memory, computational, energy, and latency requirements may exceed the
resources available at the robotic edge. These constraints motivate lighter
routing mechanisms that can execute locally, while preserving as much
selection quality as possible.

This paper studies tool routing as a bounded multi-label selection problem. The router cannot invent commands or generate arbitrary executable code. Instead, it must select a subset from a fixed allowlisted catalog of 42 robot tools. We compare four implementations of this routing function: CDBv1, which uses a fixed semantic-distance threshold; CDBv2, which uses a threshold relative to the nearest result; CDBv3, which decomposes compound prompts into intent-like clauses and aggregates their retrieval results; and a cloud-hosted
120-billion-parameter LLM that performs direct selection from the complete
catalog.

The evaluation uses 164 manually annotated requests divided into three
complexity levels: 43 direct single-tool requests, 52 consequential requests
requiring exactly two tools, and 69 compound requests requiring three or more
tools. All routers are evaluated against the same ground-truth assignments
using exact-match accuracy, completeness, precision, a
completeness-weighted composite score, and routing latency.

The main contributions of this work are the formalisation of robotic tool routing as bounded multi-label capability selection, an empirical comparison of fixed-threshold, relative-distance, and clause-aware semantic retrieval under the same tool catalogue and prompt dataset, and a direct comparison between the strongest retrieval-only method and a large cloud-hosted LLM, exposing the resulting trade-off between routing quality and latency.


The results show that CDBv3 is the strongest retrieval-only strategy. Its
macro-average composite score is $0.7822$, compared with $0.5863$ for CDBv2
and $0.3330$ for CDBv1. The 120B LLM achieves a higher score of $0.8869$, but
CDBv3 provides substantially lower routing latency and does not require cloud
connectivity. The results therefore do not establish retrieval as a universally more accurate replacement for large-model routing. Instead, they show that clause-aware semantic retrieval is a practical local alternative when latency, connectivity, computation, privacy, or data sovereignty are primary deployment constraints.
\section{Related Work}

Tool-using language-model agents combine natural-language interpretation with calls to external functions or services. ReAct alternates reasoning, action selection, and observation, allowing an agent to update its behaviour after receiving the result of a tool invocation~\cite{yao2023react}. Toolformer
investigates whether language models can learn when to call external tools,
which tool to use, which arguments to provide, and how to incorporate the
returned result into subsequent generation~\cite{schick2023toolformer}.
ToolLLM extends this problem to large collections of real-world APIs and
includes both single-tool and multi-tool instructions~\cite{qin2024toolllm}.

In embodied systems, Code as Policies translates natural-language instructions
into programs composed from predefined perception and control
APIs~\cite{liang2023codeaspolicies}. SPINE uses language models to interpret
incomplete mission specifications and connect them to semantic planning for
ground and aerial robots~\cite{ravichandran2025spine}. These systems show that
language models can mediate between human requests and executable robotic
functions. Their principal focus, however, is task planning, policy generation,
or orchestration rather than a controlled comparison of routing mechanisms for
selecting from an allowlisted set of independently executable robot tools.

The present work treats tool selection as a separate classification and
retrieval problem. The router cannot invent arbitrary commands, modify tool
names, or directly generate shell instructions. It must return one or more
known capability definitions whose executable interfaces are specified in
advance. This restriction bounds the action space and separates probabilistic
language understanding from the robot's execution boundary.

Dense retrieval represents queries and candidate records as vectors in a
shared embedding space, allowing semantically related items to be matched even
when they do not contain the same surface terms. Dense Passage Retrieval
demonstrated the effectiveness of learned dense representations for selecting
relevant passages from large collections~\cite{karpukhin2020dense}.
Retrieval-augmented generation subsequently combined a neural retriever with a
generative model, conditioning generation on retrieved external
records~\cite{lewis2020rag}. Although the present study does not use retrieval
to provide candidates to a smaller generative model, the underlying retrieval
principle is relevant: a natural-language request is compared against textual
descriptions of known capabilities, and the nearest capability records are
used to determine the routing output.

ChromaDB (CDB) is used as the implementation substrate for dense capability
retrieval~\cite{chromadb}. Each allowlisted tool is stored with a textual
summary, description, and associated metadata. At query time, the user request
is embedded and compared with the stored capability embeddings. CDB
supports approximate nearest-neighbour search through a Hierarchical Navigable
Small World index~\cite{malkov2020hnsw}. CDB itself is not proposed as a
new routing algorithm in this work; it provides vector storage, indexing, and
similarity search. The evaluated contribution lies in the routing policies
implemented above this infrastructure, including fixed-distance filtering,
relative-distance filtering, and clause-aware aggregation.

Related retrieval-based systems have also been studied in software engineering.
Chen et al. evaluate a retrieval-augmented framework in which relevant code is
selected from a repository before being supplied to a language
model~\cite{chen2024codesearch}. Koziolek et al. retrieve validated industrial
function blocks for inclusion in LLM-generated control
programs~\cite{koziolek2024control}. These works demonstrate how retrieval can
connect generative models to existing domain-specific components. In contrast,
the present work does not generate control code. It evaluates whether semantic
retrieval alone can select the predefined processes required by a robotic
request.

Agent-oriented software systems further illustrate the value of bounded tool
sets. RepairAgent plans and invokes software-analysis and validation tools
during program repair~\cite{bouzenia2025repairagent}, while Batole et al. use
an agent-oriented approach for locating software-design
issues~\cite{batole2025localizeagent}. SALLMA proposes separate operational and
knowledge layers for LLM-based multi-agent systems, including intent
management, orchestration, and persistent agent configuration
~\cite{becattini2025sallma}. These architectures are related to the separation
between capability descriptions, routing, and execution considered here, but
their evaluation targets software repositories or broader agent architectures
rather than real-time robotic capability selection.

The reliability of the routing boundary is important because retrieval and LLM
components introduce integration failure modes beyond model accuracy alone. A
study of open-source LLM applications with retrieval support identified
recurring functionality, efficiency, and security defects
~\cite{shao2025integration}. These findings support treating routing, output
validation, and process execution as explicit components rather than combining
them into an unrestricted end-to-end agent.

Previous research has established that LLMs can invoke tools, generate
policies, coordinate multi-step tasks, and use retrieved context. Less attention has been given to the narrower routing boundary between an unconstrained natural-language request and a fixed set of robot-facing processes. In particular, there is limited empirical evidence comparing different embedding-only retrieval policies with direct large-model selection under the same capability catalog, annotations, and evaluation metrics. This work addresses that gap by comparing three progressively refined retrieval methods with a 120B LLM baseline, considering both routing quality and latency.

\section{Problem Formulation}
\label{sec:problem_formulation}

Let
\begin{equation}
\mathcal{T}=\{t_1,t_2,\ldots,t_n\}
\label{eq:tool_catalog}
\end{equation}
denote the finite catalog of $n$ tools available to the robotic system. Each
tool $t_i\in\mathcal{T}$ is identified by a unique name and is associated with
a natural-language summary, a detailed description, an executable interface,
and usage metadata. The executable interface is defined in advance; the router
is not allowed to create new tools or arbitrary commands.

An input request $x\in\mathcal{X}$ is a natural-language prompt submitted by a
user and may express one or more intents related to the robot's capabilities. A tool router performs multi-label selection: it may return no tool, one tool, or several tools. The routing function is therefore defined as
\begin{equation}
f:\mathcal{X}\rightarrow 2^{\mathcal{T}},
\label{eq:routing_function}
\end{equation}
where $2^{\mathcal{T}}$ denotes the power set of the tool catalog.
For a request $x$, let $Y(x)\subseteq\mathcal{T}$ denote the ground-truth set
of required tools and let
\begin{equation}
\widehat{Y}(x)=f(x)
\end{equation}
denote the set predicted by the router. For example, the request
\texttt{``Tell me the battery level.''} requires only the battery-monitoring
tool, yielding
$Y(x)=\{\texttt{battery}\}$. In contrast, the request
\texttt{``If the battery is below 20\% or the wind exceeds 10 m/s, land.''}
requires two observation tools and one actuation tool:
\begin{equation}
Y(x)=\{\texttt{battery},\texttt{wind\_speed},\texttt{land}\}.
\end{equation}
Returning only $\widehat{Y}(x)=\{\texttt{land}\}$ is incomplete because the
system cannot evaluate either triggering condition. Returning
$\widehat{Y}(x)=\{\texttt{battery},\texttt{wind\_speed},\texttt{land},\linebreak[2]
\texttt{whereami}\}$ is complete, but includes one unnecessary tool.

Router performance is evaluated using three request-level measures:
exact-match accuracy $A(x)$, completeness $C(x)$, and precision $P(x)$.
Exact-match accuracy is one only when the predicted and reference sets are
identical:
\begin{equation}
A(x)=\mathds{1}\left[\widehat{Y}(x)=Y(x)\right].
\label{eq:prompt_accuracy}
\end{equation}

Completeness is a binary request-level measure indicating whether all required
tools are present, regardless of additional selections:
\begin{equation}
C(x)=\mathds{1}\left[Y(x)\subseteq\widehat{Y}(x)\right].
\label{eq:prompt_completeness}
\end{equation}
This measure differs from conventional multilabel recall because partial
coverage of the required set still gives $C(x)=0$.

Precision penalises unnecessary selections and is defined by
\begin{equation}
P(x)=
\begin{cases}
\dfrac{|\widehat{Y}(x)\cap Y(x)|}{|\widehat{Y}(x)|},
& |\widehat{Y}(x)|>0,\\[10pt]
1, & \widehat{Y}(x)=Y(x)=\varnothing,\\[3pt]
0, & \widehat{Y}(x)=\varnothing\ \land\ Y(x)\neq\varnothing.
\end{cases}
\label{eq:prompt_precision}
\end{equation}

Let
\begin{equation}
\mathcal{D}=\{(x_j,Y_j)\}_{j=1}^{m},
\qquad Y_j=Y(x_j),
\label{eq:dataset}
\end{equation}
denote an annotated dataset containing $m$ requests. Dataset-level exact-match
accuracy, completeness, and precision are macro-averaged over requests:
\begin{align}
\mathcal{A} &= \frac{1}{m}\sum_{j=1}^{m}A(x_j),
\label{eq:dataset_accuracy}\\
\mathcal{C} &= \frac{1}{m}\sum_{j=1}^{m}C(x_j),
\label{eq:dataset_completeness}\\
\mathcal{P} &= \frac{1}{m}\sum_{j=1}^{m}P(x_j).
\label{eq:dataset_precision}
\end{align}

Because omitting a required sensing or action tool may prevent the request from being executed, completeness is given greater weight in the composite metric. For each router and request type, we define
\begin{equation}
\mathcal{L}
=
0.25\,\mathcal{A}
+
0.75
\frac{5\,\mathcal{C}\mathcal{P}}
{4\,\mathcal{P}+\mathcal{C}}.
\label{eq:router_score}
\end{equation}
The second term is an $F_2$-style harmonic combination of completeness and
precision. Exact-match accuracy remains a separate component to reward
predictions that are both complete and minimal. The composite score is treated
as a summary measure; the individual values of $\mathcal{A}$, $\mathcal{C}$,
and $\mathcal{P}$ are also reported because different deployments may assign
different costs to false negatives and false positives.

The formulation evaluates selection only. It does not determine tool arguments, execution order, control flow, or whether logical and temporal dependencies are preserved after routing. Those concerns belong to downstream planning and orchestration and are outside the scope of this study.
\section{Routing Strategies}
\label{sec:routing_strategies}

We evaluate four implementations of the routing function in
\equref{routing_function}. All methods receive the same natural-language
request and select from the same catalog $\mathcal{T}$. CDBv1, CDBv2, and
CDBv3 are retrieval-only methods implemented with CDB. 
The fourth method
is a direct large-model baseline.

\subsubsection{CDBv1: Fixed-Threshold Retrieval}

At startup, each tool representation is embedded and inserted into an in-memory
ChromaDB collection. At runtime, the complete user request is embedded and
queried against the collection. Let $d(x,t_i)$ denote the vector distance
between request $x$ and tool $t_i$. CDBv1 returns
\begin{equation}
\widehat{Y}_{\mathrm{v1}}(x)
=
\{t_i\in\mathcal{T}:d(x,t_i)\leq\tau_{\mathrm{abs}}\},
\end{equation}
where $\tau_{\mathrm{abs}}=0.85$ is a fixed global threshold.

This method is inexpensive and straightforward to inspect. However, a single
threshold must account for requests with different ambiguity and semantic
complexity. A strict value may omit required tools, particularly for compound
requests, whereas a permissive value may return several semantically related
but unnecessary tools.

\subsubsection{CDBv2: Relative-Distance Retrieval}

CDBv2 replaces the absolute threshold with a rule relative to the nearest
result. For request $x$, let
\begin{equation}
d_{\min}(x)=\min_{t_i\in\mathcal{T}}d(x,t_i).
\end{equation}
The nearest tool is always retained, and additional tools are selected when
\begin{equation}
d(x,t_i)\leq(1+\rho)d_{\min}(x),
\qquad \rho=0.60.
\end{equation}

The relative rule adapts the candidate-set size to the local distance
structure of each query. When several tools are similarly close to the
request, several may be returned. When the nearest result is clearly separated
from the remaining catalog, the method returns a smaller set. CDBv2 therefore
reduces dependence on one absolute distance scale, but it still embeds each
request as a single undivided sequence.

\subsubsection{CDBv3: Clause-Aware Aggregated Retrieval}

CDBv3 is designed to improve retrieval for requests containing several
conditions or actions. It searches both the complete request and a set of
intent-like fragments extracted from it. The query set includes: (i) the full
prompt; (ii) fragments separated by commas, semicolons, and coordinating terms
such as \texttt{and} and \texttt{then}; and (iii) condition or action fragments
around terms such as \texttt{if}, \texttt{when}, \texttt{before}, and
\texttt{after}. Duplicate or empty fragments are discarded.

For example, the request
\begin{quote}
\texttt{``if the battery is low, land and announce a warning''}
\end{quote}
can generate separate queries for \texttt{``the battery is low''},\linebreak[2]
\texttt{``land''}, and\linebreak[2] \texttt{``announce a warning''}
, in addition to the complete prompt. Each fragment is queried independently. For every fragment,
CDBv3 starts with the nearest match and retains at most three tools whose
distance remains close to the best result.

The retrieved records are then aggregated by tool. A capability that is
supported by several prompt fragments receives evidence from several searches,
while lexical matches between the request and the tool name, summary, or
description provide additional support. The aggregated candidates are ranked,
and the list is truncated using a score-gap value of \texttt{0.28} and a
maximum output size of \texttt{8}. CDBv3 remains a retrieval-only router: no
language model performs a second selection stage.

\subsubsection{Direct 120B LLM Baseline}

The baseline provides a cloud-hosted 120-billion-parameter language model with
the complete tool catalog and the original user request. The model is instructed
to return the smallest complete set of catalog tools required to satisfy the
request, using a fixed JSON schema and without inventing or renaming tools.

Unlike embedding retrieval, the LLM can explicitly interpret logical
operators, monitored conditions, dependencies, and compound instructions. For
example, it may infer that conditional landing requires both the tool that
observes the condition and the tool that performs the action. The baseline,
however, requires the complete catalog to be included in every request and
depends on cloud connectivity, structured-output reliability, and remote
inference latency.

\section{Routing Results}
\label{sec:routing_results}

\subsection{Experimental Setup}
\label{sec:experimental_methodology}

The 42-tool catalog (sensors and actuators) was designed by a robotics
domain expert, who ensured that tool capabilities do not overlap. The
prompt set was constructed to cover every sensor and actuator at each
complexity level, maximizing diversity, and the same expert assigned the
ground-truth tool set to each prompt.

CDB methods use \texttt{mxbai-embed-large:v1} (Ollama) with ChromaDB
\texttt{1.5.9} (distance metric \texttt{l2}, index configuration
$M{=}16$,\linebreak[2]
\texttt{ef\_construction}{=}100,\linebreak[2]
\texttt{ef\_search}{=}100).
The CDBv1--CDBv3 hyperparameters
(Section~\ref{sec:routing_strategies}) were selected through manual
iteration directly against the same 164-prompt benchmark used for
evaluation, rather than on an independent development set, so the scores
reported below should be read as best-case, upper-bound estimates.
The direct LLM baseline uses \texttt{gpt-oss:120b-cloud} 
(digest \texttt{569662207105}; temperature 1.0; 120s timeout, no retry)
and is run 10 times per
prompt, in contrast to the deterministic CDB routers, which are run once.
Malformed responses are parsed as JSON with a regex fallback; out-of-catalog
names are kept as false positives; empty or unparseable responses yield no
selected tools.

Latency measurements were performed on an Intel Core i5-9600K CPU \linebreak[2] (3.70GHz),
24GB DDR4 RAM, and an NVIDIA RTX 5060 Ti (16GB VRAM) used for local inference
acceleration, running Ubuntu.
The cloud baseline was accessed over a home broadband network via a wired Ethernet connection.

The analysis is organised in two stages. 
First, CDBv1, CDBv2, and CDBv3 are
compared to determine how the successive retrieval refinements affect routing quality. 
Second, the strongest retrieval-only method, CDBv3, is compared with the direct 120B LLM baseline. Results are segmented by request type and reported using exact-match accuracy, completeness, precision, the composite score $\mathcal{L}$, and end-to-end routing latency.

\subsection{CDBv1, CDBv2 and CDBv3}

\begin{figure}
    \centering
    \includegraphics[width=0.95\linewidth]{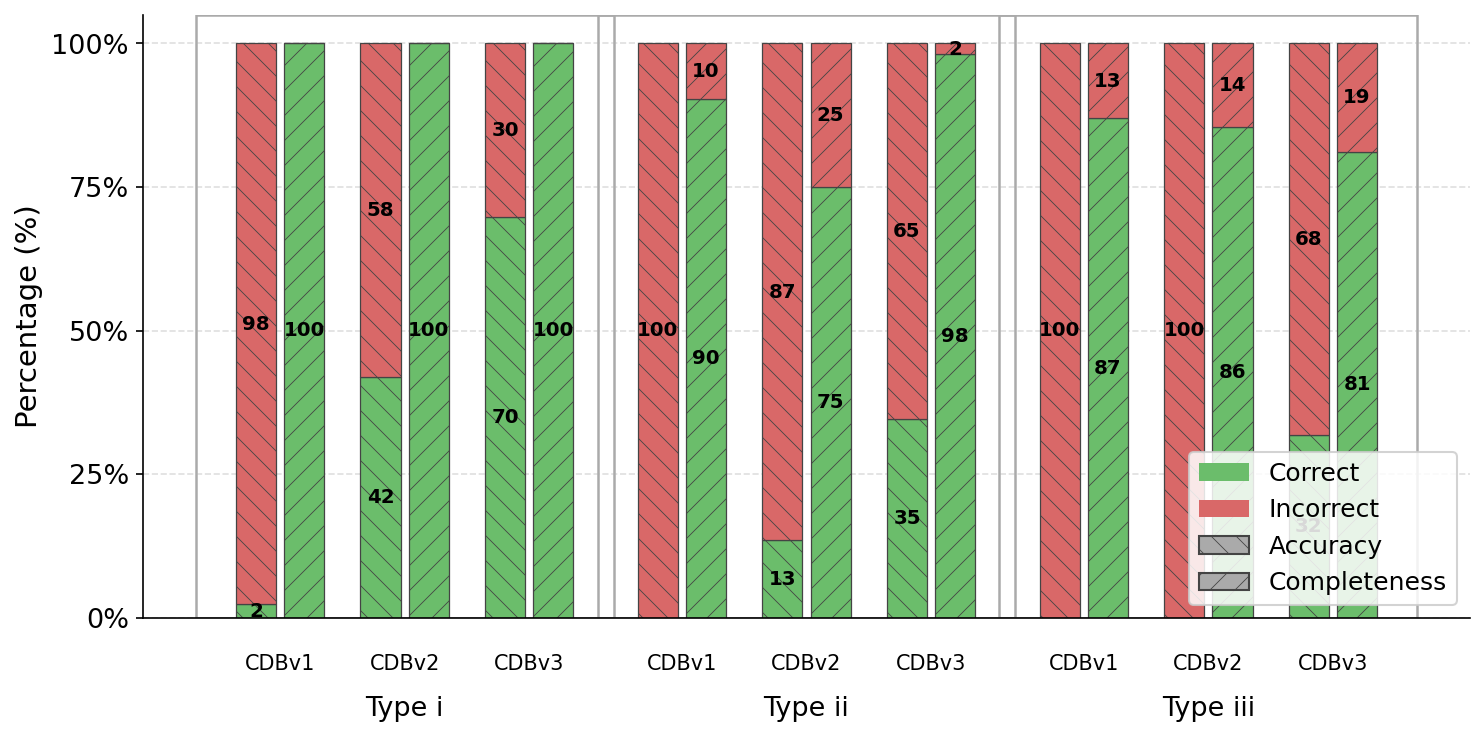}
    \vspace{-0.3cm}
    \caption{Exact-match accuracy and completeness of CDBv1, CDBv2, and CDBv3, segmented by request type.}
    \label{fig:A_C}
\end{figure}

Figure~\ref{fig:A_C} shows that the retrieval refinements substantially improve performance over the fixed-threshold baseline. CDBv1 has very low exact-match accuracy: its fixed threshold keeps most required tools (completeness between 87\% and 100\%) but also admits many unrelated ones. CDBv2 improves direct single-tool routing by adapting the cutoff to the nearest result, but its performance remains limited for compound requests.
CDBv3 provides the strongest overall retrieval-only result. Clause-level
queries allow different conditions and actions to contribute separate evidence, which improves completeness for two-tool requests and accuracy for multi-tool requests.
The improvement is not uniform across every metric and request. CDBv3 may
retrieve additional semantically related tools, so gains in completeness can be accompanied by false positives. Precision must therefore be considered together with exact match and completeness rather than treating retrieval recall as the only objective.

\begin{figure}[!ht]
    \centering
    \includegraphics[width=0.90\linewidth]{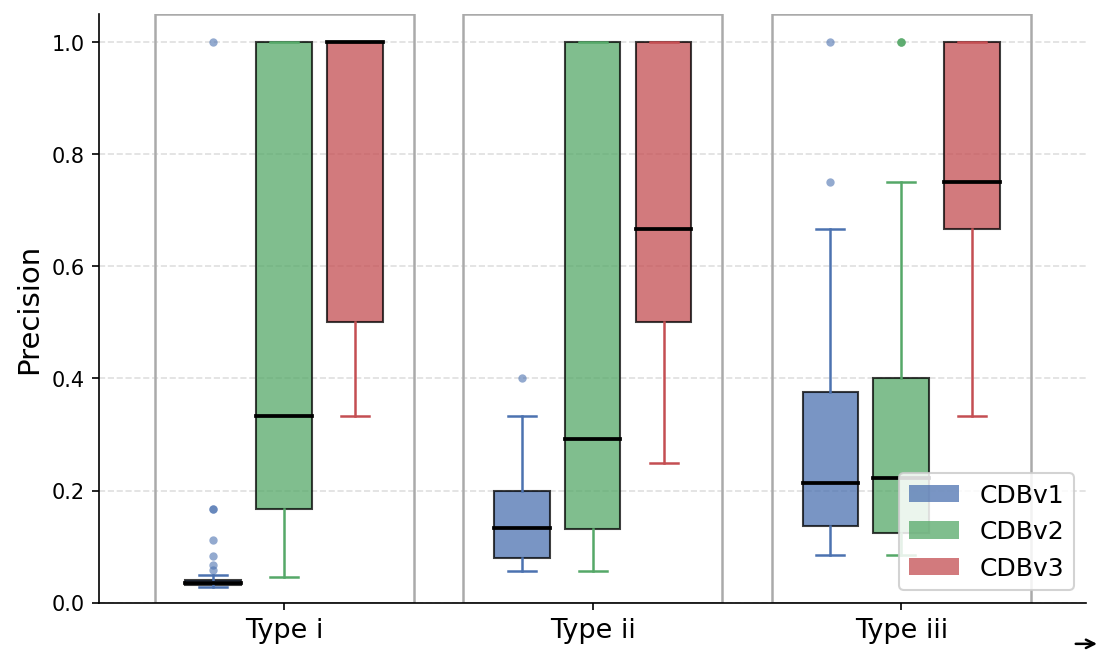}
    \vspace{-0.3cm}
    \caption{Request-level precision of CDBv1, CDBv2, and CDBv3, segmented by request type.}
    \label{fig:P}
\end{figure}

Figure~\ref{fig:P} confirms that CDBv1's permissive behaviour does not translate into a useful overall router: although it rarely omits required tools, it returns many unnecessary ones, and its precision distribution remains poor. CDBv3 achieves the highest composite score among the retrieval-only methods because it maintains high completeness while substantially improving exact-match accuracy and precision, particularly on compound requests.

\subsection{CDBv3 versus the 120B LLM Baseline}

\begin{figure}[!ht]
    \centering
    \includegraphics[width=0.93\linewidth]{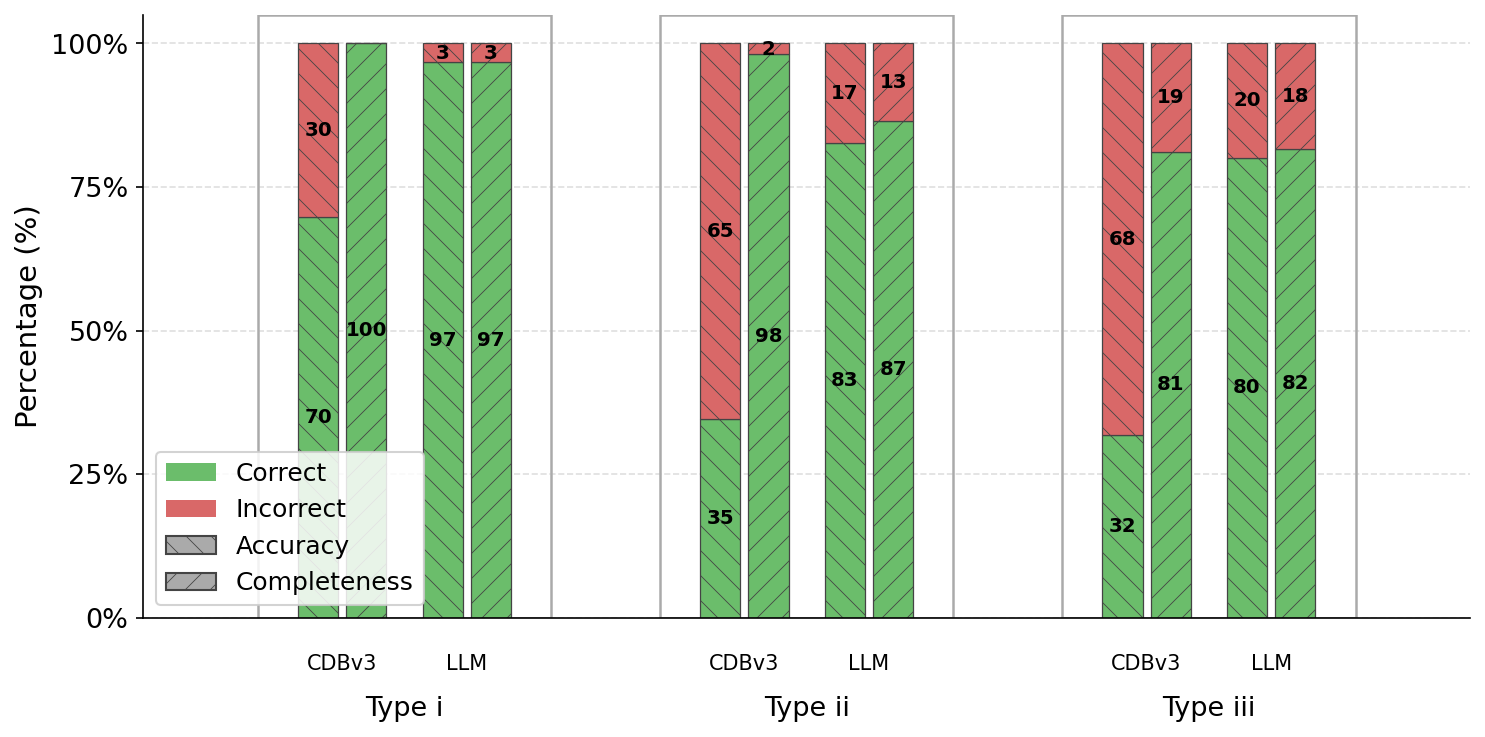}
    \vspace{-0.3cm}
    \caption{Exact-match accuracy and completeness of CDBv3 and the direct 120B LLM baseline, segmented by request type.}
    \label{fig:A_C2}
\end{figure}

The direct LLM baseline outperforms CDBv3 in exact-match accuracy
across the three request types, as shown in Fig.~\ref{fig:A_C2}, whereas
completeness is similar or higher for CDBv3. The LLM's advantage therefore
comes mainly from returning fewer unnecessary tools rather than from covering
more required ones.
The difference is smallest for direct single-tool
requests and becomes more relevant when the prompt requires several conditions or actions to be interpreted jointly. This result is consistent with the LLM's ability to reason explicitly over logical relations while viewing the complete tool catalog.

\begin{figure}[!ht]
    \centering
    \includegraphics[width=0.80\linewidth]{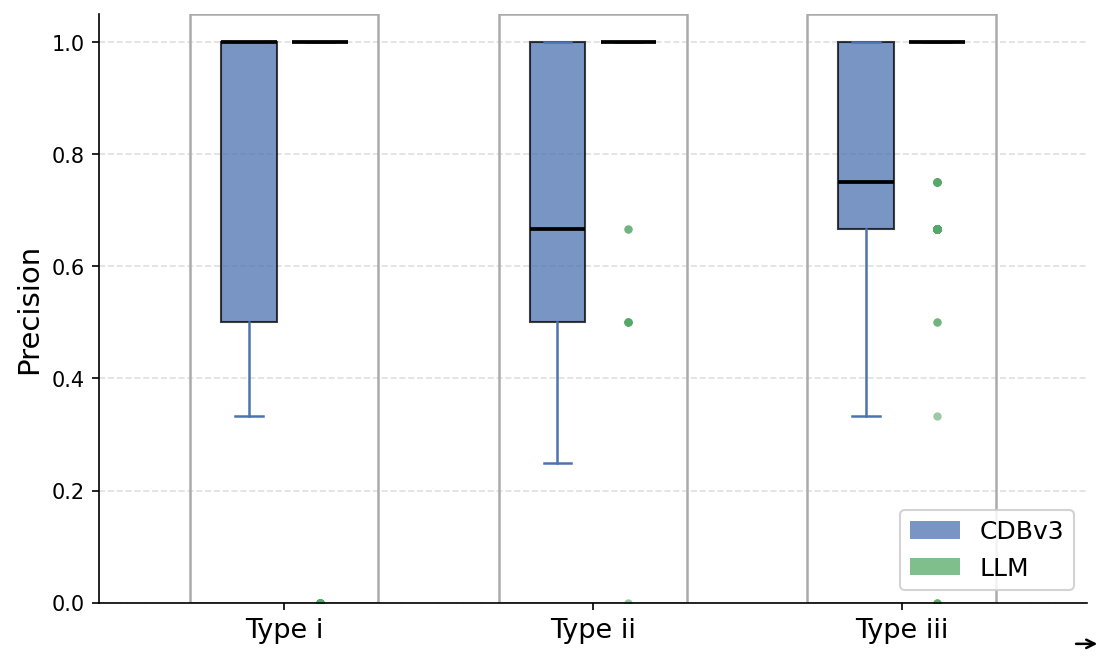}
    \vspace{-0.3cm}
    \caption{Request-level precision of CDBv3 and the direct 120B LLM baseline, segmented by request type.}
    \label{fig:P2}
\end{figure}

The precision distributions in Fig.~\ref{fig:P2} show that both methods can produce minimal outputs for many requests. Nevertheless, CDBv3 has a broader distribution for more complex prompts because clause-aware retrieval may retain several related candidates even when only a subset is required. The LLM more frequently removes these semantically plausible but unnecessary tools.

\begin{figure}[!ht]
    \centering
    \includegraphics[width=0.8\linewidth]{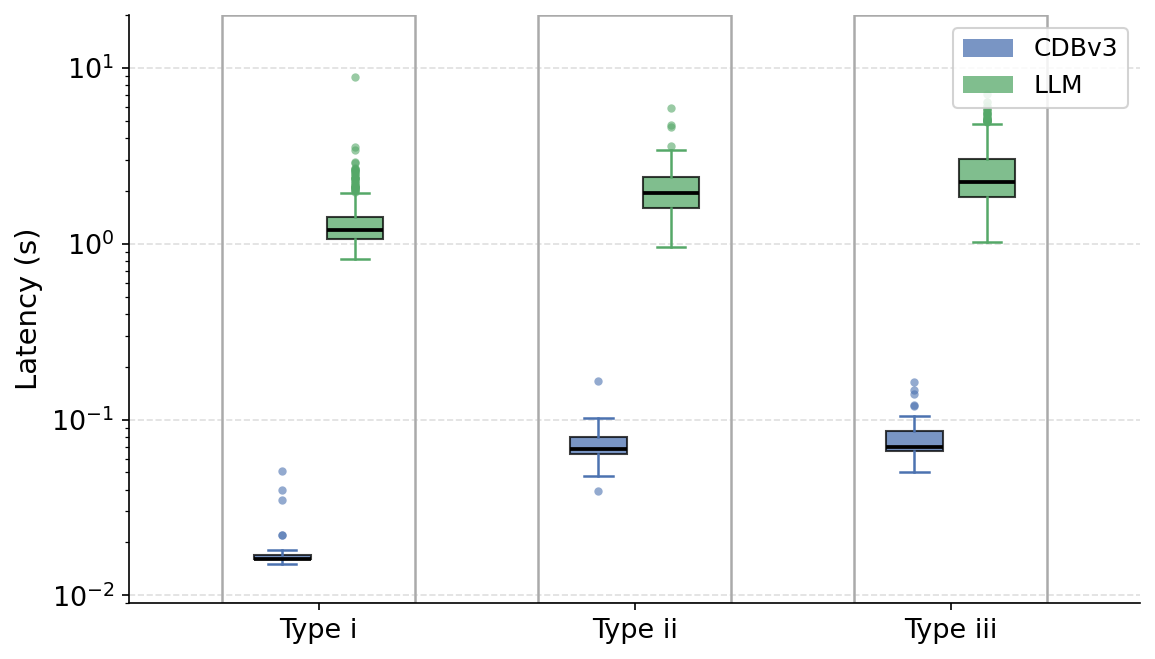}
    \caption{End-to-end routing latency of CDBv3 and the cloud-hosted 120B LLM baseline, segmented by request type.}
    \label{fig:router_timing_comparison}
    \vspace{-0.3cm}
\end{figure}

Latency is measured end-to-end, so it includes not only algorithmic execution
but also confounding factors such as network conditions, cloud server
availability, and inference queuing. We report it here in a comparative,
order-of-magnitude sense rather than as an isolated measure of computation:
even accounting for these factors, the local system consistently operates on
the order of milliseconds, while the cloud-hosted baseline operates on the
order of seconds.

The quality advantage of the LLM is accompanied by substantially higher and
more variable latency. Figure~\ref{fig:router_timing_comparison} shows that
CDBv3 remains close to the local retrieval time across all request types,
whereas the cloud baseline exhibits second-scale latency and larger outliers.
The comparison therefore reveals a deployment trade-off rather than a single dominant method: the LLM provides better routing quality, while CDBv3 provides faster and connectivity-independent selection.

\subsection{Composite Score}

Table~\ref{tab:router_l_scores} reports the composite score for each request type. 
The final column is the macro-average of the three type-level
scores.
CDBv3 is the strongest retrieval-only router for all three types and improves the macro-average score by approximately $0.196$ over CDBv2. 
The direct LLM retains the highest routing score, exceeding CDBv3 by approximately $0.105$.
Accordingly, the principal result of the short study is not that retrieval
surpasses the large model in routing quality, but that clause-aware retrieval closes part of the quality gap while preserving a markedly lighter and faster deployment profile.

\begin{table}
\centering
\caption{Composite score $\mathcal{L}$ for each router and request type.}
\label{tab:router_l_scores}
\begin{tabular}{lcccc}
\toprule
\textbf{Router} &
\textbf{Type~i} &
\textbf{Type~ii} &
\textbf{Type~iii} &
\textbf{Macro mean} \\
\midrule
CDBv1 & 0.2058 & 0.3385 & 0.4548 & 0.3330 \\
CDBv2 & 0.7462 & 0.5380 & 0.4748 & 0.5863 \\
CDBv3 & 0.8932 & 0.7723 & 0.6812 & 0.7822 \\
LLM   & 0.9674 & 0.8636 & 0.8298 & 0.8869 \\
\bottomrule
\end{tabular}
\end{table}
\vspace{-0.5cm}
\section{Conclusion}
\label{sec:conclusion}

This paper investigated the mapping of natural-language requests to the
complete and minimal set of predefined robot tools required to satisfy them.
The problem was formulated as bounded multi-label capability selection and
evaluated using three embedding-based ChromaDB routers and a direct
cloud-hosted 120-billion-parameter LLM baseline. The short benchmark contains
164 manually annotated requests divided into direct single-tool, two-tool
consequential, and three-or-more-tool compound categories.

CDBv3 achieved the strongest performance among the retrieval-only methods. Its
macro-average composite score was $0.7822$, compared with $0.5863$ for CDBv2
and $0.3330$ for CDBv1. Clause-aware prompt decomposition and aggregation
therefore provide a substantial improvement over both fixed-threshold and
single-query relative-distance retrieval. The direct LLM baseline nevertheless
achieved the highest routing quality, with a macro-average score of $0.8869$.
This distinction is important: the results support CDBv3 as a practical local
alternative, but they do not show that retrieval-only routing is more accurate
than direct large-model selection.

The latency results expose a complementary advantage. CDBv3 executes locally
and produces substantially lower and less variable end-to-end routing latency,
whereas the baseline depends on cloud inference and network conditions. The
preferred routing strategy therefore depends on deployment priorities. Direct
large-model routing is preferable when selection quality is the dominant
criterion and cloud access is acceptable. Clause-aware semantic retrieval is
attractive when local execution, predictable response time, privacy,
connectivity independence, or limited edge resources are more important.


The paper has a limitation regarding 
 retrieval-only results: the CDBv1--CDBv3 hyperparameters (Section~\ref{sec:experimental_methodology})
were tuned through manual iteration on the same 164-prompt benchmark used for
the final evaluation, rather than on a held-out development set. The reported
CDBv1--CDBv3 scores should therefore be read as best-case estimates rather
than as an evaluation on independent data.
We also tested a small, locally deployable LLM (7-8B class) as a candidate
edge baseline; the results were discouraging, with routing quality well
below both CDBv3 and the 120B model, so it was excluded from the main
comparison.
%
Future work should evaluate larger and evolving catalogs, separate development
and test sets for routing-parameter selection, robustness to paraphrases and
prompt injection, and execution-aware metrics. The use of a smaller language
model to refine a retrieval-generated candidate set is intentionally left for a separate study.

\section*{Acknowledgments}
This work was supported by FCT -- Fundação para a Ciência e Tecnologia, I.P. under the project Agent-Based Cooperative Autonomy for Multi-UAV Systems, ref. LASIGE-SP-2026.06, under the LASIGE Research Unit, ref. UID/00408/2025, DOI https://doi.org/10.54499/UID/00408/2025 \hfil
and  2023.15441.TENURE.051/CP00003/CT00029.

\bibliographystyle{splncs04}
\bibliography{refs}

@inproceedings{yao2023react,
  author    = {Yao, Shunyu and Zhao, Jeffrey and Yu, Dian and Du, Nan and Shafran, Izhak and Narasimhan, Karthik R. and Cao, Yuan},
  title     = {{ReAct}: Synergizing Reasoning and Acting in Language Models},
  booktitle = {Proceedings of the 11th International Conference on Learning Representations},
  year      = {2023}
}

@inproceedings{schick2023toolformer,
  author    = {Schick, Timo and Dwivedi-Yu, Jane and Dess{\`i}, Roberto and Raileanu, Roberta and Lomeli, Maria and Hambro, Eric and Zettlemoyer, Luke and Cancedda, Nicola and Scialom, Thomas},
  title     = {Toolformer: Language Models Can Teach Themselves to Use Tools},
  booktitle = {Advances in Neural Information Processing Systems},
  year      = {2023},
  volume    = {36},
  pages     = {68539--68551},
  publisher = {Curran Associates, Inc.}
}

@inproceedings{qin2024toolllm,
  author    = {Qin, Yujia and Liang, Shihao and Ye, Yining and Zhu, Kunlun and Yan, Lan and Lu, Yaxi and Lin, Yankai and Cong, Xin and Tang, Xiangru and Qian, Bill and Zhao, Sihan and Hong, Lauren and Tian, Runchu and Xie, Ruobing and Zhou, Jie and Gerstein, Mark and Li, Dahai and Liu, Zhiyuan and Sun, Maosong},
  title     = {{ToolLLM}: Facilitating Large Language Models to Master 16000+ Real-World {APIs}},
  booktitle = {Proceedings of the 12th International Conference on Learning Representations},
  year      = {2024}
}

@inproceedings{liang2023codeaspolicies,
  author    = {Liang, Jacky and Huang, Wenlong and Xia, Fei and Xu, Peng and Hausman, Karol and Ichter, Brian and Florence, Pete and Zeng, Andy},
  title     = {Code as Policies: Language Model Programs for Embodied Control},
  booktitle = {2023 IEEE International Conference on Robotics and Automation},
  year      = {2023},
  pages     = {9493--9500},
  publisher = {IEEE}
}

@inproceedings{ravichandran2025spine,
  author    = {Ravichandran, Zachary and Murali, Varun and Tzes, Mariliza and Pappas, George J. and Kumar, Vijay},
  title     = {{SPINE}: Online Semantic Planning for Missions with Incomplete Natural Language Specifications in Unstructured Environments},
  booktitle = {2025 IEEE International Conference on Robotics and Automation},
  year      = {2025},
  pages     = {13714--13721},
  publisher = {IEEE}
}

@inproceedings{karpukhin2020dense,
  author    = {Karpukhin, Vladimir and O{\u{g}}uz, Barlas and Min, Sewon and Lewis, Patrick and Wu, Ledell and Edunov, Sergey and Chen, Danqi and Yih, Wen-tau},
  title     = {Dense Passage Retrieval for Open-Domain Question Answering},
  booktitle = {Proceedings of the 2020 Conference on Empirical Methods in Natural Language Processing},
  year      = {2020},
  pages     = {6769--6781},
  publisher = {Association for Computational Linguistics}
}

@inproceedings{lewis2020rag,
  author    = {Lewis, Patrick and Perez, Ethan and Piktus, Aleksandra and Petroni, Fabio and Karpukhin, Vladimir and Goyal, Naman and K{\"u}ttler, Heinrich and Lewis, Mike and Yih, Wen-tau and Rockt{\"a}schel, Tim and Riedel, Sebastian and Kiela, Douwe},
  title     = {Retrieval-Augmented Generation for Knowledge-Intensive {NLP} Tasks},
  booktitle = {Advances in Neural Information Processing Systems},
  year      = {2020},
  volume    = {33},
  pages     = {9459--9474},
  publisher = {Curran Associates, Inc.}
}

@misc{chromadb,
  author       = {{Chroma}},
  title        = {{ChromaDB} Documentation},
  year         = {2026},
  howpublished = {Online documentation},
  note         = {Accessed: 2026-06-24}
}

@article{malkov2020hnsw,
  author  = {Malkov, Yu. A. and Yashunin, D. A.},
  title   = {Efficient and Robust Approximate Nearest Neighbor Search Using Hierarchical Navigable Small World Graphs},
  journal = {IEEE Transactions on Pattern Analysis and Machine Intelligence},
  year    = {2020},
  volume  = {42},
  number  = {4},
  pages   = {824--836}
}

@inproceedings{chen2024codesearch,
  author    = {Chen, Junkai and Hu, Xing and Li, Zhenhao and Gao, Cuiyun and Xia, Xin and Lo, David},
  title     = {Code Search Is All You Need? Improving Code Suggestions with Code Search},
  booktitle = {Proceedings of the IEEE/ACM 46th International Conference on Software Engineering},
  year      = {2024},
  articleno = {73},
  numpages  = {13},
  publisher = {Association for Computing Machinery}
}

@inproceedings{koziolek2024control,
  author    = {Koziolek, Heiko and Gr{\"u}ner, Sten and Hark, Rhaban and Ashiwal, Virendra and Linsbauer, Sofia and Eskandani, Nafise},
  title     = {{LLM}-Based and Retrieval-Augmented Control Code Generation},
  booktitle = {Proceedings of the 1st International Workshop on Large Language Models for Code},
  year      = {2024},
  pages     = {22--29},
  publisher = {Association for Computing Machinery}
}

@inproceedings{bouzenia2025repairagent,
  author    = {Bouzenia, Islem and Devanbu, Premkumar T. and Pradel, Michael},
  title     = {{RepairAgent}: An Autonomous, {LLM}-Based Agent for Program Repair},
  booktitle = {2025 IEEE/ACM 47th International Conference on Software Engineering},
  year      = {2025},
  pages     = {2188--2200},
  publisher = {IEEE}
}

@inproceedings{batole2025localizeagent,
  author    = {Batole, Fraol and O'Brien, David and Nguyen, Tien N. and Dyer, Robert and Rajan, Hridesh},
  title     = {An {LLM}-Based Agent-Oriented Approach for Automated Code Design Issue Localization},
  booktitle = {2025 IEEE/ACM 47th International Conference on Software Engineering},
  year      = {2025},
  pages     = {1320--1332},
  publisher = {IEEE}
}

@inproceedings{becattini2025sallma,
  author    = {Becattini, Marco and Verdecchia, Roberto and Vicario, Enrico},
  title     = {{SALLMA}: A Software Architecture for {LLM}-Based Multi-Agent Systems},
  booktitle = {2025 IEEE/ACM International Workshop on New Trends in Software Architecture},
  year      = {2025},
  pages     = {5--8},
  publisher = {IEEE}
}

@inproceedings{shao2025integration,
  author    = {Shao, Yuchen and Huang, Yuheng and Shen, Jiawei and Ma, Lei and Su, Ting and Wan, Chengcheng},
  title     = {Are {LLMs} Correctly Integrated into Software Systems?},
  booktitle = {2025 IEEE/ACM 47th International Conference on Software Engineering},
  year      = {2025},
  pages     = {1178--1190},
  publisher = {IEEE}
}

\end{document}